\pdfoutput=1

\documentclass[11pt]{article}

\usepackage[]{ACL2023}

\usepackage{times}
\usepackage{latexsym}
\usepackage{amsmath}
\usepackage{graphicx}
\usepackage[T1]{fontenc}
\definecolor{average}{HTML}{006101}
\definecolor{ideal}{HTML}{286ba8}
\usepackage{enumitem}
\usepackage{tabularx}
\usepackage{booktabs}  
\usepackage{multirow}
\usepackage{enumitem} 

\usepackage[utf8]{inputenc}

\usepackage{microtype}

\usepackage{inconsolata}

\usepackage{natbib} 
\title{VoiceAgentEval: A Dual-Dimensional Benchmark for Expert-Level Intelligent Voice-Agent Evaluation of Xbench's Professional-Aligned Series}

\makeatletter
\renewcommand{\thanks}[1]{%
  \protected@xdef\@thanks{\@thanks\protect\footnotetext[0]{#1}}%
}

\renewcommand\@makefntext[1]{%
  \parindent 0pt
  \noindent
  \@makefnmark #1%
}
\makeatother

\author{\small Pengyu Xu$^{1,4*}$, Shijia Li$^{1,5*}$, Ao Sun$^{1,6*}$, Feng Zhang$^{1,8}$, Yahan Li$^{1,7}$, Bo Wu$^{1,8}$\\
\small \textbf{Zhanyu Ma$^{1}$, Jiguo Li$^{1}$, Jun Xu$^{1}$, Jiuchong Gao$^{1}$, Jinghua Hao$^{1}$, Renqing He$^{1}$, Rui Wang$^{2}$}\\
\small \textbf{Yang Liu$^{2}$, Xiaobo Hu$^{2}$, Fan Yang$^{3}$, Jia Zheng$^{3}$, Guanghua Yao$^{3}$}\\
\small$^1$ Meituan \qquad 
\small$^2$ Xbench \qquad 
\small$^3$ Agora \qquad 
\small$^4$ Beijing Jiaotong University \qquad 
\small$^5$ BUPT \qquad \\
\small$^6$ The Chinese University of Hong Kong, Shenzhen \qquad 
\small$^7$ Jilin University \qquad 
\small$^8$ Peking University 
\thanks{$*$ Equal Contribution \\
\textsuperscript{1}Our paper's webpage: \url{https://github.com/LVYUERLVR/OutboundEval-Xbench}
}
}

\begin{document}
\maketitle
\begin{abstract}
We propose \textbf{VoiceAgentEval}, a comprehensive benchmark for evaluating large language models (LLMs) in expert-level intelligent outbound calling scenarios. Unlike existing methods that suffer from three key limitations—insufficient dataset diversity and category coverage, unrealistic user simulation, and inaccurate evaluation metrics—VoiceAgentEval addresses these issues through a structured framework. First, we design a benchmark spanning six major business domains and 30 representative sub-scenarios, each with scenario-specific process decomposition, weighted scoring, and domain-adaptive metrics. Second, we develop a large-model-driven \textbf{User Simulator} that generates diverse, persona-rich virtual users with realistic behaviors, emotional variability, and communication styles, providing a controlled yet authentic testing environment. Third, we introduce a dynamic evaluation method that adapts to task variations, integrating automated and human-in-the-loop assessment to measure task execution accuracy, professional knowledge application, adaptability, and user experience quality. Experiments on 12 state-of-the-art LLMs reveal distinct trade-offs between expert-level task completion and interaction fluency, offering practical insights for building reliable, human-like outbound AI systems. VoiceAgentEval\textsuperscript{1} establishes a practical, extensible, and domain-oriented standard for benchmarking LLMs in professional applications.

\end{abstract}

\begin{figure*}[htbp]
  \centering
  \includegraphics[width=0.95\linewidth]{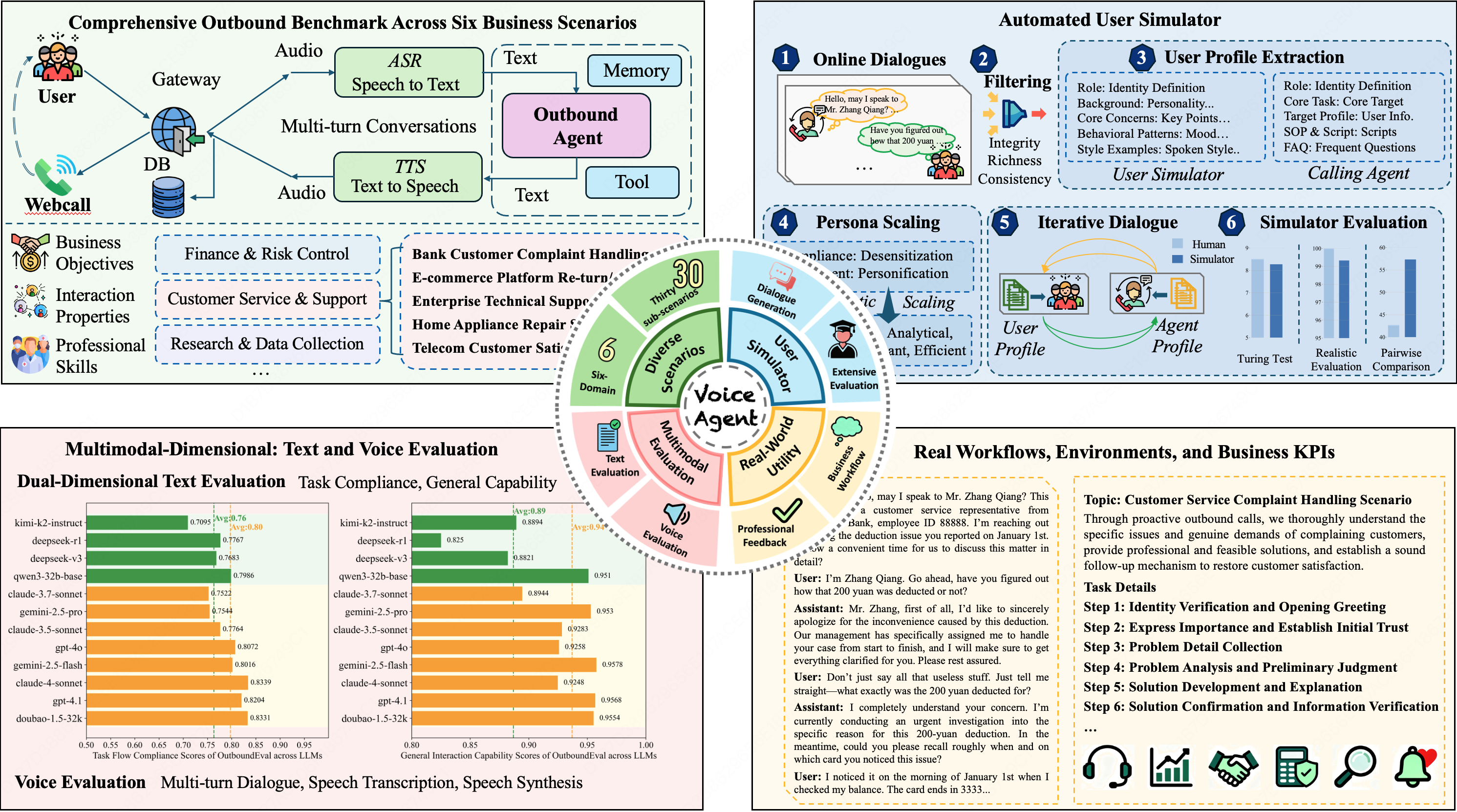}
  \caption{The VoiceAgentEval Framework for Evaluating Outbound Interaction Systems}
  \label{fig:arch1}
\end{figure*}

\section{Introduction}

With the pervasive integration of Large Language Models (LLMs) across various industries, AI-driven automated outbound calling is emerging as a critical component for enterprises to optimize customer communication and enhance operational efficiency~\citep{wen2025voice,kaewtawee2025cloningconversationalvoiceai,kaiyrbekov2025automated,lang2025telephone}. Its applications span a wide range of domains, including recruitment, market research, sales, and customer service, with several benchmarks—such as Xbench~\citep{chen2025xbench}—having established evaluation criteria for them. However, a standardized benchmark specifically designed for outbound calling scenarios is currently lacking to comprehensively and objectively evaluate the performance of these models in real-world tasks. Existing evaluation efforts predominantly focus on general conversational abilities or single-turn instruction following, and suffer from insufficient dataset volume and category coverage, unrealistic user simulation, and inaccurate or unreasonable evaluation metrics.

To address this gap, we introduce VoiceAgentEval, an evaluation framework designed to drive the advancement of outbound calling AI towards greater intelligence, human-like interaction, and efficiency. This framework assesses the capabilities of outbound calling agents from three primary dimensions: benchmark development, user simulator, and evaluation methodology. Key features of this framework include:

\begin{itemize}[leftmargin=*]
  \item \textbf{Benchmark Development:} We have constructed a comprehensive, scenario-based corpus derived from authentic outbound calling business data. This corpus encompasses six major business domains and 30 representative sub-scenarios. For each sub-scenario, we have established a detailed evaluation scheme that includes scenario-specific process decomposition, a weighted scoring system, and domain-adaptive metrics, forming a solid foundation for nuanced and objective assessment.

  \item \textbf{User Simulator:}  To facilitate scalable and consistent evaluation, we propose a systematic process for constructing user simulators. By leveraging interaction data from real-world business scenarios, we build a large number of effective and stable user simulators. This allows for the testing of models in a controlled and reproducible environment, examining their task completion capabilities across various communication styles.

  \item \textbf{Evaluation Methodology:} In terms of text evaluation, we have developed a dual-layer assessment system comprising task process adherence and general interaction capabilities, achieving precise evaluation through standardized scenario design and quantitative scoring mechanisms. For speech evaluation, we have established 15 metrics across three major scenarios, which integrate expert scoring with objective data to provide multi-dimensional quantification of speech recognition accuracy, system robustness, audio quality, and interactive experience. This comprehensive evaluation framework offers clear technical guidance for optimizing outbound AI systems.

\end{itemize}
Through this comprehensive design, VoiceAgentEval can thoroughly examine the dialogue generation capabilities and voice interaction quality of AI agents in outbound calling scenarios, offering an objective basis for performance comparison and optimization among different models.

\section{Related Work}

\subsection*{Evaluation of Task-Oriented Multi-Turn Dialogue}
Task-oriented dialogue systems aim to help users accomplish concrete goals such as booking or information lookup, which aligns with outbound calling use cases. Traditional evaluation relies on static, pre-collected corpora and thus cannot faithfully reflect robustness and policy effectiveness under dynamic interaction~\citep{chen2021action,andreas2020task,budzianowski2018multiwoz}. ComplexBench measures instruction following with multi-constraint compositions at scale, but its finely grained constraint combinatorics are not well matched to the behaviors required in dialogue~\citep{wen2024benchmarking}. This motivates evaluation protocols that capture both global task success and local linguistic competence over multi-turn speech interactions. 

This work conducts a comprehensive evaluation of task completion and linguistic capability in outbound calling scenarios, considering both the overall dialogue process and its local segments.

\subsection*{Benchmarks for Speech Language Models}

AIR-Bench evaluates SLM understanding of human speech, natural sounds, and music, yet treats audio primarily as contextual input while questions and instructions remain textual~\citep{yang2024air}. SD-Eval targets paralinguistic understanding such as emotion, accent, environment, and age, and VoiceBench assesses LLM-based spoken dialogue in realistic scenarios; however, both report only textual outputs and therefore omit evaluation of speech output quality, which is critical in spoken agents~\citep{ao2024sd,chen2024voicebench}. ADU-Bench measures SLM performance on open-domain audio dialogue understanding but does not include multi-turn interaction~\citep{gao2024benchmarking}. URO-Bench proposes a comprehensive suite for end-to-end spoken dialogue, addressing gaps in multilingual ability, multi-turn interaction, and certain non-linguistic dimensions~\citep{yan2025uro}. However, it lacks sufficient focus and guidance for practical business scenarios.

This work takes real business scenarios of SLMs as the foundation, integrating text-based evaluation of task completion with speech-based assessment of interaction quality. By doing so, it provides a comprehensive examination of AI agents’ dialogue generation and spoken interaction in outbound calling contexts, offering an objective basis for cross-model comparison and optimization.

\subsection*{User Simulation with Language Models}
Language-model-driven simulators have been used to generate interactive agents across settings, including non-player characters in text games~\citep{kim2022plm}, multi-agent social environments~\citep{wu2024autogen,park2023generative}, and human–AI interaction for online shopping or web search~\citep{chen2024chatshop,zhang2024usimagent}. $\tau$-bench is the first to deploy LM role simulators for automated agent reliability testing, focusing on retail and airline customer service and demonstrating the feasibility and value of simulation-based evaluation~\citep{yao2024tau}. Yet prior simulators rarely couple dialogue evaluation with spoken output assessment, and few are grounded in outbound calling workflows.

This work leverages real data from a broader range of business scenarios to construct a large set of simulated human roles. These role simulations create a controllable and reproducible environment for testing agents, enabling systematic evaluation of their performance under different communication styles.

\section{Benchmark Development}
We implement multi-scenario generalization assessment through a carefully designed domain system and evaluation strategy.

\paragraph{Six-Domain Scenario System}
We define six major outbound domains to ensure coverage and representativeness:

\begin{enumerate}[leftmargin=*]
    \item \textbf{Customer Service and Support} — Problem resolution with emotional sensitivity and technical accuracy. These scenarios uniquely require \textit{emotion repair orientation}, where AI must simultaneously resolve issues and transform adversarial customer emotions into collaborative interactions.
    \vspace{-0.3cm}
    \item \textbf{Sales and Marketing} — Conversion-oriented persuasion and objection handling. Distinguished by \textit{defense breakthrough orientation}, requiring data-driven value demonstration to overcome customers' natural skepticism and defensive attitudes toward commercial motives.
    \item \textbf{Human Resources Management} — Information gathering and candidate matching in a professional yet friendly manner. Characterized by \textit{bidirectional evaluation orientation}, where AI must both assess candidates and be evaluated by them while handling career-sensitive information.
    \vspace{-0.3cm}
    \item \textbf{Finance and Risk Control} — High-precision, compliance-sensitive interactions involving sensitive data. Uniquely demands \textit{compliance constraint orientation} with zero-tolerance for errors while balancing strict regulatory requirements with customer experience.
    \vspace{-0.3cm}
    \item \textbf{Research and Data Collection} — Structured elicitation of information through guided questioning. Requires \textit{neutrality maintenance orientation}, establishing research authenticity while avoiding any perception of commercial bias.
    \vspace{-0.3cm}
    \item \textbf{Proactive Care and Notifications} — Informational outreach with a human touch. Emphasizes \textit{warmth delivery orientation}, prioritizing genuine emotional connection and brand care while addressing concerns about hidden commercial intentions.
\end{enumerate}

\begin{figure*}[htbp]
  \centering
  \includegraphics[width=0.8\linewidth]{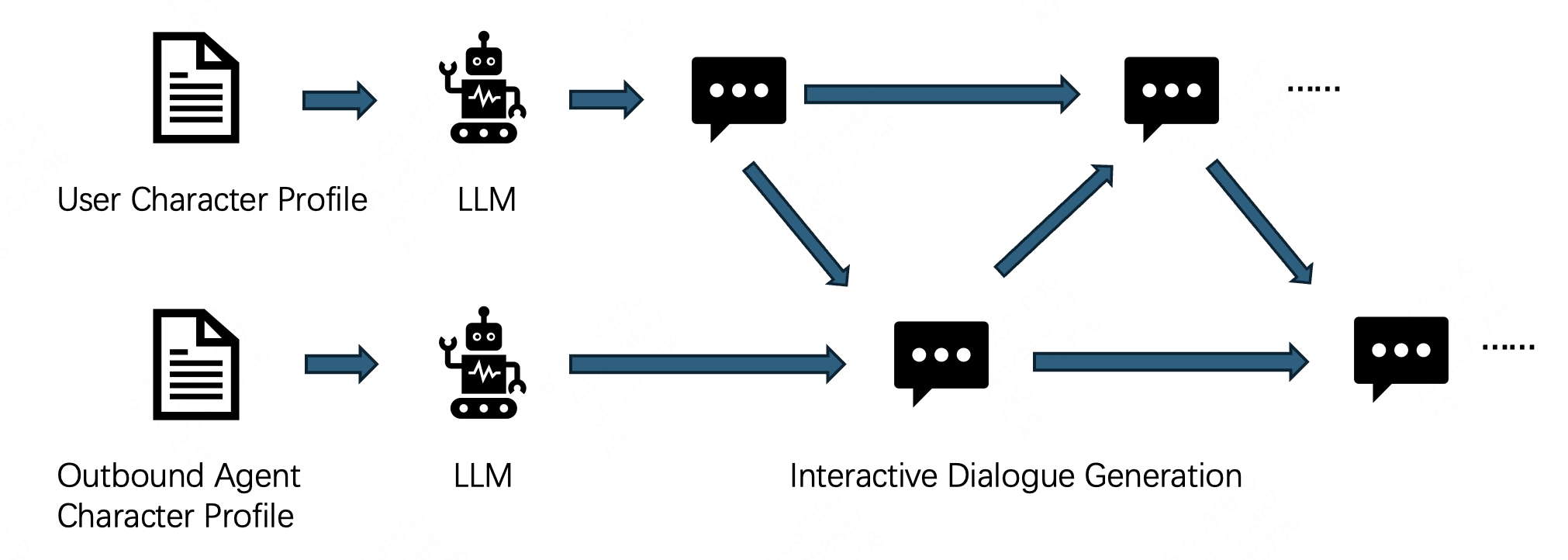}
  \caption{Interactive dialogue generation.}
  \label{fig:arch}
\end{figure*}

\begin{figure*}[htbp]
  \centering
  \includegraphics[width=0.8\linewidth]{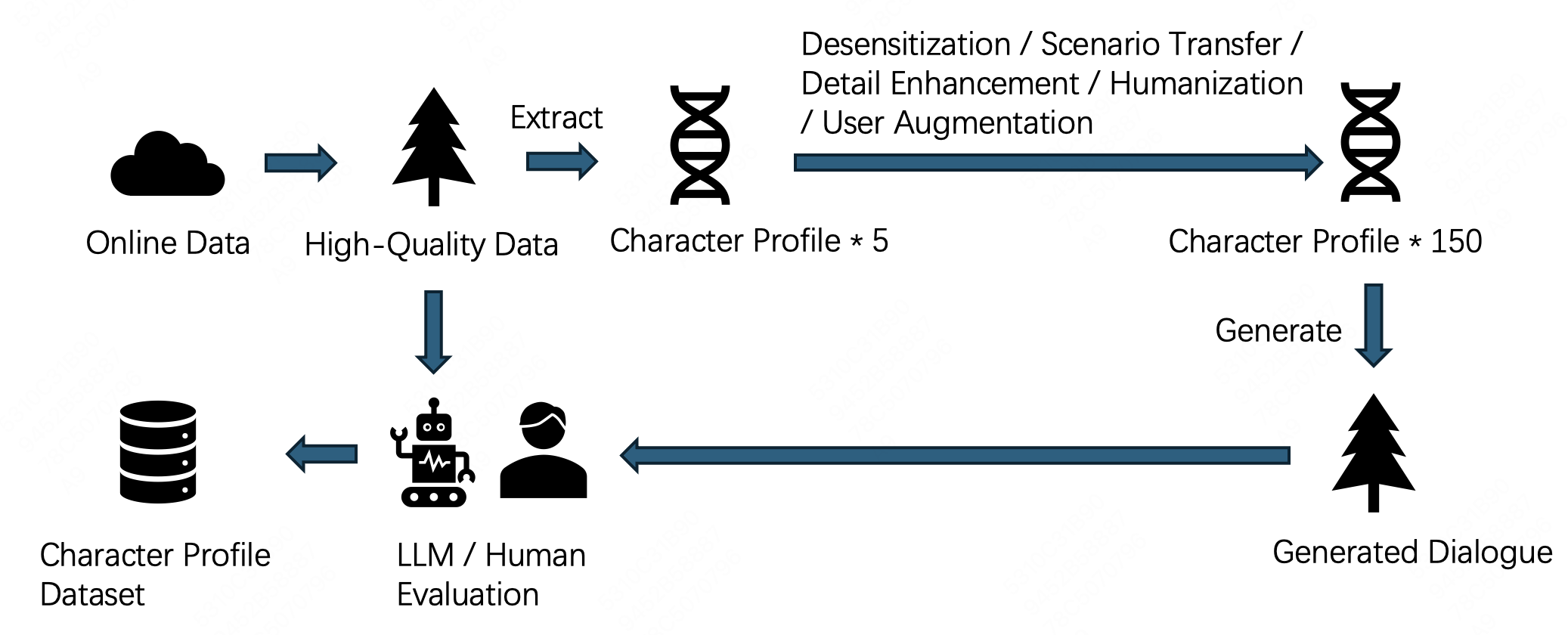}
  \caption{Generation of character profiles.}  
  \label{fig:arch2}
\end{figure*}

VoiceAgentEval encompasses the current mainstream outbound call scenarios. The data is categorized into six major directions. Each major direction contains five sub-scenarios, resulting in a total of 30 sub-scenarios. Furthermore, each sub-scenario includes five different user types, leading to a total of 150 data entries. A detailed description of the data is presented in the table~\ref{table:customer_service_and_support_scenarios}~\ref{table:sales_and_marketing_scenarios}~\ref{table:human_resource_management_scenarios}~\ref{table:finance_and_risk_control_scenarios}~\ref{table:research_and_information_collection_scenarios}~\ref{table:proactive_care_notifications_scenarios} in Appendix.



\section{User Simulator}

\subsection{User Simulator Design}
In the domain of automated outbound calling services, evaluating an AI agent's ability to complete tasks in realistic scenarios is a core yet challenging problem. The primary difficulties arise from the complexity of real-world users, the stochastic nature of communication, and the volatility of human emotions. These factors make it difficult for traditional script-based testing methods to objectively and comprehensively assess the robustness and adaptability of AI agents.  
To address this bottleneck, we introduce the concept of a \textbf{User Simulator}.

Our proposed simulator is a \textit{prompt-based} large language model (LLM), currently implemented using the GPT-4.1~\citep{openaigpt41blog} architecture (with the potential for iterative upgrades). The simulator plays the role of a diverse set of phone call recipients, engaging with the AI agent under evaluation in dynamic, real-time voice or text interactions. By configuring varied behavioral patterns, knowledge backgrounds, and dialogue strategies, we create a controlled and reproducible experimental environment. This enables systematic evaluation of the agent's task completion, adaptability, and communication skills when interacting with different user personalities. Such an approach not only improves evaluation efficiency and scalability but also provides standardized benchmarks for iterative optimization.

\subsection{Definition}
The User Simulator is a prompt-based LLM. During agent evaluation, we employ two distinct prompt-based LLMs: one as the \emph{User}, and the other as the \emph{Outbound Agent}. The purpose of this interactive setup is to assign each simulated user persona with unique behavior patterns, background knowledge, and conversational strategies, allowing the AI simulator to act as the recipient. This setup ensures that agent performance can be evaluated in a reproducible environment across varying communication styles.

\subsection{Construction Pipeline}
We design a systematic pipeline comprising data preparation, persona generation, humanization enhancement, and large-scale expansion to construct a high-fidelity user simulator. The pipeline ensures logical coherence and high behavioral realism. The specific process is shown in Figure \ref{fig:arch} and Figure \ref{fig:arch2}.

\subsubsection{Structured Persona Definition}
Simulator behavior is driven by a structured \emph{persona} document, divided into two parts:  \\
\textbf{Calling Agent (Outbound Agent)}:
\begin{itemize}[leftmargin=*]
    
    \item \textbf{Role}: Definition of the agent's identity.
    \vspace{-0.3cm}
    \item \textbf{Core Task}: A concise, single-sentence description of the primary goal.
    \vspace{-0.3cm}
    \item \textbf{Target Profile}: Basic pre-known information about the user.
    \vspace{-0.3cm}
    \item \textbf{SOP \& Script}: Standard operating procedures and recommended dialogue.
    \vspace{-0.3cm}
    \item \textbf{FAQ}: Standard responses to common user questions.
    \vspace{-0.3cm}
    \item \textbf{Opening}: Standardized call introduction.
    \vspace{-0.3cm}
    \item \textbf{Termination Conditions}: Criteria for successful or failed task completion and corresponding closing statements.
\end{itemize}

\noindent \textbf{User Simulator (Receiving Party)}:
\begin{itemize}[leftmargin=*]
    \item \textbf{Role}: Definition of the user's identity.
    \vspace{-0.3cm}
    \item \textbf{Background}: Demographics, key experiences, personality traits, and pre-call context.
    \vspace{-0.3cm}
    \item \textbf{Core Concerns}: Ranked list of primary, secondary, and latent user concerns.
    \vspace{-0.3cm}
    \item \textbf{Behavioral Patterns}: Descriptions of emotional shifts, uncertainty, and environmental distractions.
    \vspace{-0.3cm}
    \item \textbf{Style Exemplars}: Positive and negative examples of conversational style to avoid ``AI-like'' tone.
    \vspace{-0.3cm}
    \item \textbf{Termination Conditions}: Triggers for ending the call from the user's side.
\end{itemize}

\subsubsection{Pipeline Steps}
\paragraph{Step 1: Seed Data Curation}  
We begin with real-world online conversation data, scoring them on multi-dimensional quality metrics (completeness, logical consistency, informativeness). The top 5 dialogues are selected as seed data for persona modeling.

\paragraph{Step 2: Initial Persona Extraction}  
From the 5 seed samples, we manually craft 5 structured personas following the schema.

\paragraph{Step 3: Scenario Generalization}  
To protect business-sensitive information, personas are recontextualized to generalized outbound domains (e.g., customer service, HR management, finance) using LLM rewriting while preserving information structure and granularity.

\paragraph{Step 4: Data De-identification}  
We systematically replace personal, corporate, and sensitive details using strict substitution rules (e.g., standardized names, placeholder companies, fixed reference dates).

\paragraph{Step 5: Detail Enrichment}  
We enhance SOPs and user backgrounds to improve realism and completeness.

\paragraph{Step 6: Humanization Enhancement}  
We add personality depth via:
\begin{itemize}[leftmargin=*]
    \item \textbf{Persona Enrichment}: Adding situational context affecting dialogue tone.
    \item \textbf{Human Fallibility}: Incorporating emotional variability, memory lapses, biases, and noise.
    \item \textbf{Core Concerns Definition}: Identifying explicit, implicit, and emotional needs.
    \item \textbf{Vernacular Shaping}: Calibrating colloquial style using positive/negative examples.
\end{itemize}

\paragraph{Step 7: Persona Scaling}  
We expand the 30 finalized personas to 150 using a standardized personality archetype matrix:
\begin{enumerate}[leftmargin=*]
    \item \textbf{Cooperative}: Communicative and positive, with clear needs, friendly demeanor, and prompt decision-making.
    \item \textbf{Analytical}: Neutral in communication, reliant on information, emotionally calm, and makes decisions after analysis.
    \item \textbf{Hesitant}: Passive in communication, unclear needs, repeatedly seeks confirmation, and slow in decision-making.
    \item \textbf{Resistant}: Avoids communication, lacks clear needs, shows impatience, and tends to reject directly.
    \item \textbf{Efficient}: Goal-oriented with a focus on efficiency; emotionally stable and pursues quick results.

\end{enumerate}
This ensures diversity and coverage across user types.

\subsection{Dialogue Generation and Evaluation}
Using the 150 persona pairs, we generate simulated dialogues between the User Simulator and the target AI agent.  

\begin{table}[h]
\centering
\begin{tabular}{|l|c|}
\hline
 & \textbf{Score (out of 9)} \\ \hline
Human Dialogue & \textbf{8.50} \\ \hline
AI-Generated Dialogue & 8.27 \\ \hline
\end{tabular}
\caption{Score comparison between human and AI-generated dialogues.}
\end{table}

\begin{table}[h]
\centering
\begin{tabular}{|l|c|c|}
\hline
\textbf{Realistic Evaluation} & \textbf{Human} & \textbf{AI} \\ \hline
Human Dialogue & 150 & 0 \\ \hline
AI Dialogue & 149 & 1 \\ \hline
\end{tabular}
\caption{Realistic evaluation of human and AI dialogues.}
\end{table}

\begin{table}[h]
\centering
\begin{tabular}{|l|c|c|c|}
\hline
\textbf{Pairwise} & \textbf{Total} & \textbf{Human Win} & \textbf{AI Win} \\ \hline
Count & 150 & 64 & 86 \\ \hline
Percentage & 100\% & 42.67\% & 57.33\% \\ \hline
\end{tabular}
\caption{Pairwise comparison results between human and AI-generated dialogues.}
\end{table}

We adopt three evaluation strategies:
\begin{enumerate}[leftmargin=*]
    \item \textbf{Human-likeness Scoring}: A Turing-test-inspired 0--9 scale based on linguistic naturalness, imperfection, and emotional plausibility.
    \item \textbf{AI Detection}: Binary classification by an LLM to determine whether a user is AI-generated.
    \item \textbf{Pairwise Comparison}: Side-by-side judgment of real vs.\ simulated dialogues.
\end{enumerate}

\subsubsection{Human Evaluation}

We conduct blind reviews with human annotators:
\begin{itemize}[leftmargin=*]
    \item \textbf{AI Detection}: Classifying each dialogue as human- or AI-generated.
    \item \textbf{Pairwise Comparison}: Identifying which of two dialogues is more likely AI-generated.
\end{itemize}
Where applicable, statistical hypothesis testing is applied to validate results.

\section{Evaluation Methodology}

Our evaluation framework comprises two primary components: \textit{text evaluation} and \textit{speech evaluation}.  
Each task is assessed along two dimensions---\textbf{Task Flow Compliance} and \textbf{General Interaction Capability}---with weighted aggregation to obtain a comprehensive score.

\subsection{Text Evaluation}
This section focuses on evaluating the textual aspects of outbound calling performance. The text evaluation protocol in VoiceAgentEval is derived from real-world business scenarios and emphasizes instruction-following performance in goal-oriented, multi-turn dialogues.

\subsubsection{Evaluation Objectives and Scope}
The proposed benchmark is designed to comprehensively assess both the \textit{task execution capabilities} and the \textit{interaction quality} of large language models (LLMs) in realistic outbound business settings. The key objectives are:

\begin{enumerate}[leftmargin=*]
    \item \textbf{Task Execution Capability}: Measure the model's ability to accomplish predefined business goals within complex, multi-turn dialogues. Unlike single-turn Q\&A benchmarks, outbound scenarios require continuous progression of the task flow until the business objective is fulfilled, with emphasis on end-to-end task success and adherence to business processes.
    \vspace{-0.3cm}
    \item \textbf{Interaction Experience Capability}: Assess the naturalness and coherence of dialogue, the appropriateness of emotional responses, and robustness in handling incomplete or noisy information. This dimension focuses on fundamental conversational skills that directly impact user perception and satisfaction in outbound calling contexts.
    \vspace{-0.3cm}
    \item \textbf{Multi-Scenario Generalization Capability}: Evaluate the model's ability to maintain consistent performance across diverse outbound business domains. By testing across 6 major business categories and 30 sub-scenarios, this dimension assesses whether the model can generalize effectively to different types of outbound scenarios rather than excelling only in specific domains, thereby measuring the model's robustness and adaptability in real-world deployment.
\end{enumerate}

\subsubsection{Dual-Layer Evaluation Framework}
To comprehensively assess both \textit{task execution capabilities} and \textit{interaction experience capabilities}, we design a dual-layer evaluation framework that reflects the dual nature of outbound calling as both a goal-driven task execution process and a user experience–driven communication activity. This framework comprises two primary dimensions: \textit{Task Flow Compliance} (TFC) and \textit{General Interaction Capability} (GIC).

\paragraph{TFC Assessment}
This dimension evaluates the model's understanding and execution accuracy of domain-specific business processes through scenario-specific evaluation criteria. Each scenario possesses a unique set of task flow evaluation dimensions derived through systematic workflow decomposition, ensuring that evaluation criteria precisely reflect the specific business requirements and procedural characteristics of each domain.

The evaluation dimension generation process follows a multi-stage pipeline: First, we input the task flow description for each scenario and perform workflow structure analysis, transforming lengthy task flow descriptions into structured workflow node information. Second, these workflow nodes are converted into assessable natural language evaluation criteria through iterative refinement. The process incorporates multiple validation checks, including coverage analysis between derived nodes and original task flows, and overlap detection among different evaluation dimensions. When coverage is insufficient or overlap is excessive, automatic repair mechanisms are triggered to optimize the dimension set. Finally, we establish prerequisite dependencies for each evaluation dimension based on workflow node relationships, enabling assessment of both task execution completion and procedural correctness.

During evaluation, the complete dialogue session serves as the assessment unit. An LLM-based evaluator scores the session against the derived dimension list, determining both task coverage and procedural adherence throughout the conversation.

\paragraph{GIC Assessment}
This dimension measures fundamental conversational competencies that transcend specific business domains, utilizing eight carefully derived evaluation dimensions that comprehensively capture the essential aspects of professional outbound communication. These dimensions emerge from Meituan's extensive real-world business experience across diverse outbound scenarios, providing a robust foundation for interaction quality assessment.

The eight evaluation dimensions comprise: \textit{naturalness}, \textit{coherence}, \textit{hallucination handling}, \textit{redundancy}, \textit{emotional richness}, \textit{intent understanding}, \textit{noise resistance}, and \textit{safety issues}. These dimensions comprehensively capture conversational authenticity, logical consistency, factual accuracy, information efficiency, emotional intelligence, user comprehension, robustness, and professional conduct respectively.

These dimensions were systematically derived through a bidirectional convergence approach: top-down analysis starting from theoretical model capabilities and progressively refining into specific interaction competencies, and bottom-up synthesis beginning with extensive bad case analysis from real deployment scenarios to identify recurring failure patterns. This dual-perspective methodology ensures comprehensive coverage of both aspirational interaction goals and practical deployment challenges encountered in Meituan's rich business ecosystem.

The assessment framework employs single-turn response evaluation as the fundamental scoring unit. This granular approach addresses the inherent limitations of LLM-based evaluation systems, where scoring multiple aspects simultaneously can lead to capability dilution and reduced assessment accuracy. By focusing evaluation attention on individual responses, we enhance scoring precision and enable more reliable capability measurement across the eight dimensions. The detailed scoring criteria and evaluation rubrics for each dimension are provided in the appendix.

\paragraph{Cross-Domain Evaluation Strategy}
These distinct orientations ensure that models must demonstrate fundamentally different capabilities across scenarios. Each domain features 5 sub-scenarios with 5 user personality types, creating 150 evaluation instances for comprehensive assessment.

\subsubsection{Scoring and Weighting Design}
To ensure comprehensive and accurate assessment, we design a systematic scoring framework that integrates both evaluation dimensions through carefully calibrated weighting schemes.

\paragraph{Score Components Definition}
Our evaluation framework generates three types of scores:

\textbf{Coverage Score (\(\mathrm{Cov}\))} combines two sub-components: \textit{Key Node Completion Rate} evaluates performance on critical workflow checkpoints that determine task success or failure; \textit{Overall Node Completion Rate} provides comprehensive assessment across the complete evaluation checklist, including branch nodes for exceptional scenarios.

\textbf{Process Score (\(\mathrm{Proc}\))} examines the correctness of task execution sequence, ensuring proper business logic adherence and preventing procedural violations.

\textbf{GIC Score} aggregates eight conversational competency dimensions with specific weights, measuring fundamental interaction capabilities that transcend business domains.

\paragraph{Scoring Formulas and Weight Allocation}
All weight assignments result from extensive empirical validation through systematic experiments comparing automated scoring consistency with human expert evaluations.

\begin{table*}[htbp]
\centering

\begin{tabularx}{\textwidth}{XXX}
\toprule
\textbf{Main Dimension} & \textbf{Sub Dimension} & \textbf{Weight} \\
\midrule
\multirow{2}{*}{Coverage Score} & Key Node Completion Rate & 0.7 \\
 & Overall Node Completion Rate & 0.3 \\
Process Score & Execution Order Score & 1 \\
\bottomrule
\end{tabularx}
\caption{Dimension and Weight Distribution}
\label{tab:dimension_weights}
\end{table*}

\begin{table}[htbp]
\centering
\begin{tabular}{lc}
\toprule
\textbf{Dimension} & \textbf{Weight} \\
\midrule
Naturalness & 0.20 \\
Coherence & 0.20 \\
Hallucination Handling & 0.20 \\
Redundancy & 0.10 \\
Emotional Richness & 0.10 \\
Intent Understanding & 0.10 \\
Noise Resistance & 0.06 \\
Safety Issues & 0.04 \\
\bottomrule
\end{tabular}
\caption{GIC Evaluation Dimensions and Weights}
\label{tab:gic_dimension_weights}
\end{table}

\noindent\textbf{TFC Score Calculation:}
\[
\mathrm{TFC}=0.70\,\mathrm{Cov}+0.30\,\mathrm{Proc}.
\]

\noindent\textbf{Final Score Calculation:}
\[
\mathrm{Final}=0.55\,\mathrm{TFC}+0.45\,\mathrm{GIC}.
\]

The Cov weighting (0.7 for key nodes, 0.3 for overall nodes) reflects the fundamental principle that successful task completion depends primarily on executing critical business steps, while comprehensive coverage provides additional robustness assessment. Furthermore, the TFC weighting (0.7 for Cov, 0.3 for Proc) prioritizes task completion degree over procedural correctness, recognizing that in outbound calling scenarios, achieving the business objective is more critical than strict adherence to execution order. Both weight allocations emerged from extensive empirical validation studies, with the 7:3 ratios demonstrating that key node performance and task coverage serve as the strongest predictors of real-world task success, while procedural adherence and overall coverage contribute valuable supplementary insights for service quality optimization.

For GIC, the eight-dimension weighting scheme prioritizes three foundational conversational competencies—\textbf{naturalness}, \textbf{coherence}, and \textbf{hallucination handling}—each weighted at 20\%, collectively accounting for 60\% of the GIC score. This reflects their critical importance in professional outbound communication. The remaining five dimensions are weighted according to their frequency of impact and business criticality in real deployment scenarios, as detailed in Table~\ref{tab:gic_dimension_weights}.

\subsubsection{Benchmark Results}

\begin{table*}[h]
\centering
\begin{tabular}{
    >{\raggedright\arraybackslash}p{0.8cm}  
    >{\raggedright\arraybackslash}p{6.3cm} 
    >{\centering\arraybackslash}p{2.5cm}  
    >{\centering\arraybackslash}p{2.5cm}    
    >{\centering\arraybackslash}p{2.7cm}  
}
\toprule
\textbf{Rank} & \textbf{Model Name} & \textbf{Total Score} & \textbf{Task Flow Compliance Score} & \textbf{General Interaction Capability Score} \\
\midrule
1 & doubao-1.5-32k~\citep{bytedancedoubao15} & \textbf{0.8881} & 0.8331 & 0.9554 \\
2 & gpt-4.1~\citep{openaigpt41blog} & 0.8818 & 0.8204 & \textbf{0.9568} \\
3 & claude-4-sonnet~\citep{anthropicclaude4} & 0.8748 & \textbf{0.8339} & 0.9248 \\
4 & gemini-2.5-flash~\citep{comanici2025gemini25pushingfrontier} & 0.8719 & 0.8016 & \textbf{0.9578} \\
5 & qwen3-32b-base~\citep{yang2025qwen3technicalreport} & 0.8672 & 0.7986 & 0.9510 \\
6 & gpt-4o~\citep{openaigpt4oblog} & 0.8606 & 0.8072 & 0.9258 \\
7 & claude-3.5-sonnet~\citep{anthropicclaude35} & 0.8448 & 0.7764 & 0.9283 \\
8 & gemini-2.5-pro~\citep{comanici2025gemini25pushingfrontier} & 0.8438 & 0.7544 & 0.9530 \\
9 & deepseek-v3~\citep{deepseekai2025deepseekv3technicalreport} & 0.8195 & 0.7683 & 0.8821 \\
10 & claude-3.7-sonnet~\citep{anthropicclaude37} & 0.8162 & 0.7522 & 0.8944 \\
11 & deepseek-r1~\citep{deepseekai2025deepseekr1incentivizingreasoningcapability} & 0.7984 & 0.7767 & 0.8250 \\
12 & kimi-k2-instruct~\citep{kimiteam2025kimik2openagentic} & 0.7904 & 0.7095 & 0.8894 \\
\bottomrule
\end{tabular}
\caption{Ranking of models by total score, task flow compliance score, and general interaction capability score.}
\label{tab:model_ranking}
\end{table*}



We conducted a comprehensive evaluation of 12 state-of-the-art large language models (LLMs) under the proposed framework. The results (Table~\ref{tab:model_ranking}) show that \textbf{doubao-1.5-32k}~\citep{bytedancedoubao15} ranked first with an overall score of 0.8881, demonstrating strong capabilities in both TFC (0.8331) and GIC (0.9554). In contrast, \textbf{kimi-k2-instruct}~\citep{kimiteam2025kimik2openagentic} ranked last due to its relatively low TFC score (0.7095), indicating that despite achieving reasonable conversational fluency, it exhibits significant deficiencies in task-oriented competencies such as complex instruction comprehension and procedural adherence.

The evaluation results further reveal several key insights:

First, model parameter count is not the sole determinant of performance. For instance, the relatively smaller \textbf{Gemini-2.5-Flash}~\citep{comanici2025gemini25pushingfrontier} slightly outperformed its larger counterpart, \textbf{Gemini-2.5-Pro}~\citep{comanici2025gemini25pushingfrontier}, in GIC; while \textbf{claude-4-sonnet}~\citep{anthropicclaude4} achieved the highest TFC score (0.8339) yet ranked only third overall due to its gap in GIC. This indicates that for specific scenarios, factors such as model architecture, training data quality, and alignment optimization strategies (e.g., RLHF) may be more critical than model scale alone.

Second, the performance of leading models shows clear stratification. The top eight models all achieved overall scores above 0.84, forming a high-performance cluster, which indicates that LLMs have reached a substantial baseline capability in outbound dialogue scenarios. However, the gap between the top-performing model and the theoretical maximum score reveals that existing technologies still face challenges in handling highly complex, multi-turn, or uncertain real-world conversations.

Third, the trade-off between TFC and GIC across models provides important insights for model selection in practical applications. For example, the “high TFC but relatively lower GIC performance” pattern (such as Claude-4-Sonnet) makes them highly suitable for standardized tasks with strict processes and low error tolerance (e.g., information verification, appointment scheduling). Conversely, models with balanced performance across both dimensions are better suited for customer service and communication scenarios that require high emotional intelligence and the ability to sustain multi-topic conversations.

In summary, current large language models have established a solid foundation for outbound scenarios, but the performance variations reflect trade-offs and different emphases among technical approaches. Future work may explore directions such as integrating the strengths of different models, developing more advanced alignment algorithms, and constructing more complex test scenarios.


\subsubsection{Human Verification}
While LLMs automatically decompose task flows and generate evaluation points, human verification by domain experts plays a critical role in ensuring framework reliability. Our human-in-the-loop validation operates across three key stages:
    
\paragraph{TFC Dimension Validation}
After LLM-based workflow decomposition, domain experts conduct comprehensive reviews focusing on dimension coverage and prerequisite dependency allocation. Human validators examine whether the derived evaluation dimensions comprehensively capture all critical business steps and verify the logical sequencing of evaluation points. Through systematic validation, we achieved over \textbf{95\% consistency} between automated decomposition and expert judgment. For cases where discrepancies occurred, human experts manually refined the dimension sets to ensure both completeness and business logic adherence, thereby establishing the rationality and comprehensiveness of our TFC evaluation framework.

\paragraph{GIC Evaluation Consistency}
For the General Interaction Capability assessment, we established systematic consistency validation between LLM-based evaluation and human expert scoring. Through iterative refinement of evaluation prompts based on human feedback, we achieved \textbf{90\% consistency} between automated and manual assessments. Cases with lower consistency triggered prompt optimization cycles, where human experts provided detailed feedback to enhance evaluation accuracy and reliability.

\paragraph{Scoring Framework Calibration}
To validate our dual-dimensional scoring approach, we conducted extensive comparative experiments between automated evaluation and human expert assessment. Multiple rounds of iterative validation across diverse scenarios enabled us to calibrate the weighting schemes and scoring formulas. This rigorous validation process confirmed the effectiveness of our current framework design, ensuring that automated scores align closely with expert professional judgment.

This comprehensive human verification process ensures that our evaluation framework maintains high accuracy, domain relevance, and practical applicability across diverse outbound calling scenarios.

\subsection{Voice Evaluation}
\subsubsection{Evaluation Objectives}
This section focuses on evaluating the speech interaction quality in outbound calling scenarios. The speech evaluation protocol in VoiceAgentEval is derived from real-world business scenarios and emphasizes the quantification of speech output naturalness, clarity, and user experience in goal-oriented, multi-turn dialogues.

\subsubsection{Metric Definition}
The intelligent customer service dialogue evaluation framework is structured around three core components: at the multi-turn dialogue level, it focuses on usability to ensure accurate processing of speech input, emphasizes interruption experience for timely detection and response to user interruptions, and aims for low response latency to minimize waiting; in speech recognition, it prioritizes transcription accuracy as the foundation for understanding and enhances robustness against accents and noisy environments; for speech synthesis, it demands high audio clarity and naturalness while pursuing fluid and authentic interactive experiences.

A detailed description is presented in the table~\ref{tab:voice_metrics}~\ref{tab:asr_metrics} in Appendix.

\section{Conclusion}
In this work, we presented \textbf{VoiceAgentEval}, a comprehensive evaluation framework for large language model-based outbound calling agents. Our work primarily contributes the following innovations: First, a real-world business data-driven evaluation framework and user simulator design; Second, a standardized pipeline with generalizable User Simulator Prompt. By integrating text-based and speech-based assessments with realistic role-simulated interactions, our approach enables fine-grained analysis of both task flow compliance and conversational quality. Experimental results demonstrate the framework's effectiveness in capturing key aspects of real-world outbound scenarios, providing valuable insights for both academic research and industrial deployment.

\section{Limitations}
While VoiceAgentEval offers a standardized and reproducible evaluation protocol, it has several limitations. First, the current role simulation is primarily prompt-based and may not fully capture the diversity and unpredictability of real human behavior. Second, the framework focuses on pre-defined outbound scenarios, which may limit generalization to novel domains. Finally, real-time factors such as network latency, speech synthesis artifacts, and ASR errors are only partially modeled, suggesting future work should incorporate more realistic end-to-end conditions.

\section*{Acknowledgment}
We thank all collaborators and domain experts for their valuable feedback during the design and evaluation of VoiceAgentEval. This work was supported in part by industry partners in the customer service and AI research communities, whose practical insights greatly enhanced the realism and applicability of the proposed framework.

\bibliography{custom}


\clearpage

\section*{Appendix}
\label{sec:appendix}

\appendix
\setcounter{section}{0}

\begin{table*}[htbp]
  \centering
  \renewcommand{\arraystretch}{1.2} 
  \begin{tabularx}{\textwidth}{|p{0.05\textwidth}|X|X|X|}
    \hline
    \textbf{Step} & \textbf{Data} & \textbf{Description} & \textbf{Operation} \\
    \hline
    1 & Online Data & Real online data & -- \\
    \hline
    2 & Seed Data & 5 pieces of high-quality real online data & Score the online data based on conversation quality to obtain 5 high-quality samples as seed data. \\
    \hline
    3 & Role Setting Data & 5 role settings for call initiators/recipients & Extract structured role settings from the seed data. \\
    \hline
    4 & Desensitized Data & 30 role settings for various scenarios & Perform scenario migration and desensitization; generalize and synthesize data for more scenarios based on seed data. \\
    \hline
    5 & Re-desensitized Data & 30 desensitized role settings & Replace placeholders and remove private/sensitive information. \\
    \hline
    6 & Detail-Enhanced Data & 30 role settings with enriched details & Add more details, especially step-by-step processes and core concerns. \\
    \hline
    7 & Anthropomorphic Data & 30 role settings with enhanced anthropomorphism & Increase human-likeness to make it hard to distinguish from real people. \\
    \hline
    8 & Expanded Data & 150 expanded role settings & Expand personas based on different user types. \\
    \hline
    9 & Evaluation (Auto) & Automatically evaluate conversation generation quality & Use LLM-as-Judger to assess generated conversation quality. \\
    \hline
    10 & Evaluation (Manual) & Manually blind-evaluate conversation generation quality & Conduct human blind evaluation to test if people can distinguish generated conversations from real ones. \\
    \hline
  \end{tabularx}
  \caption{Data processing pipeline for user simulator construction.}
  \label{tab:data_steps}
\end{table*}

\begin{table*}[htbp]
  \centering
  \renewcommand{\arraystretch}{1.2} 
  \setlength{\tabcolsep}{4pt}       
  \begin{tabularx}{\textwidth}{|p{0.20\textwidth}|p{0.40\textwidth}|X|}
            \hline
        \textbf{Sub-scenario} & \textbf{Core Objective} & \textbf{Main Challenges} \\ 
        \hline
        
        Bank customer complaint handling &
        Address abnormal account deductions, clarify deduction reasons, provide solutions, and restore customer trust. &
        Customers are dissatisfied with the bank's efficiency and lack of trust; requires building confidence within a short time while keeping promises. \\ 
        
        \hline
        
        E-commerce platform return/exchange processing &
        Assist customers in completing the return/exchange process to improve satisfaction levels. &
        Customers may have doubts about new service models; tight time constraints add pressure. \\ 
        
        \hline
        
         Enterprise technical support sales & 
         Promote enterprise-level system optimization services to customers, resolve existing technical issues, and facilitate service upgrades.& 
         Lack of trust in service providers; high budget pressures; concerns over migration risks for systems.\\ 

                 \hline
        
        Home appliance repair service scheduling &
        Confirm fault conditions for customers who have submitted repair requests and successfully arrange on-site repair appointments. &
        Customers are sensitive to costs and worry about being overcharged; requires communication in a quiet office environment with low-pressure tone. \\ 
        
        \hline
        Telecom customer satisfaction follow-up survey &
        Collect customer feedback regarding recent network issues, evaluate satisfaction levels, and gather suggestions for improvement. &
        Customers are dissatisfied with network stability; tight timelines require quick issue resolution. \\ 
        
        \hline
  \end{tabularx}
  \caption{Sub-scenario,Core Objective and Main Challenges in Customer Service and Support}
  \label{table:customer_service_and_support_scenarios}
\end{table*}

\begin{table*}[htbp]
  \centering
  \renewcommand{\arraystretch}{1.2} 
  \setlength{\tabcolsep}{4pt}       
  \begin{tabularx}{\textwidth}{|p{0.20\textwidth}|p{0.40\textwidth}|X|}
        \hline
        \textbf{Sub-scenario} & \textbf{Core Objective} & \textbf{Main Challenges} \\ 
        \hline
        Fitness Membership Sales & Promote fitness memberships and encourage customers to visit the gym for an experience. & Customers are price-sensitive and worry about not being able to persist due to heavy commitments. \\ \hline
        Cloud Service Sales & Recommend cloud service solutions to small and medium-sized manufacturing enterprises, improving their digitalization efficiency. & Customers lack trust in IT products, worrying about system complexity and data security issues. \\ \hline
        B2B Enterprise Procurement System Promotion & Attract new registered enterprise users, activate them, and guide them through core functionalities of the system to increase user engagement levels. & Customers doubt the reliability of the new platform; concerns over complex operations affecting procurement efficiency exist. \\ \hline
        Online Education Renewal Services & Encourage existing customers to renew subscriptions while increasing customer retention rates. & Customers have budget constraints and feel they lack time for continued learning after renewal fees are paid. \\ \hline
        Software Product Sales & Introduce intelligent office systems tailored for small- and medium-sized manufacturing enterprises, encouraging trials or purchase intent generation. 	& Customers harbor skepticism towards software products with a defensive mindset; they require concrete data evidence proving product value before trusting it fully.
        \\ 
        \hline
  \end{tabularx}
  \caption{Sub-scenario,Core Objective and Main Challenges in Sales and Marketing}
  \label{table:sales_and_marketing_scenarios}
\end{table*}

\begin{table*}[htbp]
  \centering
  \renewcommand{\arraystretch}{1.2} 
  \setlength{\tabcolsep}{4pt}       
  \begin{tabularx}{\textwidth}{|p{0.20\textwidth}|p{0.40\textwidth}|X|}
        \hline
        \textbf{Sub-scenario} & \textbf{Core Objective} & \textbf{Main Challenges} \\ 
        \hline
        Corporate Recruitment Outreach &
        Proactively contact job seekers for initial communication, introduce job positions and company details, and guide candidates through the interview process. &
        Candidates find it inconvenient to engage deeply in office environments; they need limited-time opportunities to establish interest in the position. \\ \hline
        
        HR Background Check &
        Emphasize verifying the authenticity of work experience, responsibilities within roles, performance at work, and reasons for leaving previous jobs. &
        Concerns about legal compliance during background checks arise; sensitive information like salary may cause discomfort or limited willingness to cooperate due to time constraints. \\ \hline
        
        Interview Scheduling for Recruitment &
        Coordinate with shortlisted candidates on confirming interview times and formats to complete scheduling agreements efficiently. &
        Candidates who are currently employed face challenges adjusting their schedules; uncertainty exists regarding interview timing arrangements. \\ \hline
        
        HR Onboarding Process Assistance &
        Assist new employees in completing onboarding tasks smoothly while ensuring proper handling of necessary formalities such as benefits processing. &
        New hires may feel apprehensive about adapting to a new environment; detailed explanations of processes can alleviate concerns over timelines and procedures.
        \\ \hline

        Internal Employee Training Promotion & 
        Encourage employees to participate in internal technical training courses aimed at improving engagement levels among staff members. &
        Employees doubt the quality of training and feel their busy work schedules create time constraints, limiting participation. \\ 
        \hline
  \end{tabularx}
  \caption{Sub-scenario,Core Objective and Main Challenges in Human Resource Management}
  \label{table:human_resource_management_scenarios}
\end{table*}

\begin{table*}[htbp]
  \centering
  \renewcommand{\arraystretch}{1.2} 
  \setlength{\tabcolsep}{4pt}       
  \begin{tabularx}{\textwidth}{|p{0.20\textwidth}|p{0.40\textwidth}|X|}
        \hline
        \textbf{Sub-scenario} & \textbf{Core Objective} & \textbf{Main Challenges} \\ 
        \hline

        Bank Credit Card Collection &
        Assist overdue customers in creating repayment plans to promote debt recovery. &
        Customers face difficulties in repaying due to decreased income and need a balance between collection effectiveness and actual affordability. \\ \hline
        
        Bank Loan Business Follow-up Verification &
        Verify the authenticity of customer credit loan application information and facilitate the approval process. &
        Customers urgently need funds but have unstable incomes, leading to concerns about approval timeframes and high expectations for quick processing. \\ \hline
        
        Bank Account Security Risk Control Services &
        Alert customers about abnormal transaction risks on their accounts and guide them toward adopting security measures. &
        Customers are highly skeptical of electronic identity verification processes; they feel unable to immediately cooperate due to work-related constraints or doubts regarding safety protocols. \\ \hline
        
        Bank Loan Collection &
        Collaborate with overdue customers on feasible installment repayment plans, reducing losses while mitigating bank risk exposure.& 
        Significant drops in income reduce repayment ability; skepticism exists around installment terms as well as potential impacts on credit scores.\\ \hline

        Financial Management Product Renewal Services &
        Encourage customers to renew or upgrade financial products upon maturity. &
        Customers prioritize liquidity needs over long-term benefits, as their risk tolerance and affordability are limited; they often prefer flexible cash flow options. 
        \\ 
        \hline
  \end{tabularx}
  \caption{Sub-scenario,Core Objective and Main Challenges in Finance and Risk Control}
  \label{table:finance_and_risk_control_scenarios}
\end{table*}

\begin{table*}[htbp]
  \centering
  \renewcommand{\arraystretch}{1.2} 
  \setlength{\tabcolsep}{4pt}       
  \begin{tabularx}{\textwidth}{|p{0.20\textwidth}|p{0.40\textwidth}|X|}
        \hline
        \textbf{Sub-scenario} & \textbf{Core Objective} & \textbf{Main Challenges} \\ 
        \hline
        Public Health Research &
        Collect data on community residents' health habits, disease prevention awareness, and healthcare service needs. &
        Customers doubt the authenticity of phone-based identity verification; tight schedules demand quick completion and limit understanding of professional health knowledge. \\ \hline
        
        Bank Customer Information Maintenance Services &
        Verify and update basic customer information for banking purposes. &
        Customers question the legitimacy of phone inquiries and worry about fraud risks during communication processes. \\ \hline
        
        Smart Home Product User Experience Research &
        Gather user feedback on smart home product usage to improve functionality based on suggestions.& 
        Users have limited time due to commuting or other constraints; they require comprehensive information collection within a short timeframe. \\ \hline

        Smart Home Market Research &
        Collect data on target users' preferences, usage patterns, and purchase intentions for smart home products. &
        Customers doubt the authenticity of research inquiries; tight schedules require quick establishment of trust and efficient information collection. \\ \hline
        
        Brand Awareness Market Research &
        Gather consumer insights on brand recognition, user experience, purchasing preferences, and suggestions for improvement. &
        Consumers may exhibit defensive attitudes toward unfamiliar calls; they need proof of research authenticity and prefer timely completion of data collection to avoid disruptions. 
        \\ 
        \hline
  \end{tabularx}
  \caption{Sub-scenario,Core Objective and Main Challenges in Research and Information Collection}
  \label{table:research_and_information_collection_scenarios}
\end{table*}

\begin{table*}[htbp]
  \centering
  \renewcommand{\arraystretch}{1.2} 
  \setlength{\tabcolsep}{4pt}       
  \begin{tabularx}{\textwidth}{|p{0.20\textwidth}|p{0.40\textwidth}|X|}
        \hline
        \textbf{Sub-scenario} & \textbf{Core Objective} & \textbf{Main Challenges} \\ 
        \hline
        Hospital Surgery Appointment Confirmation &
        Confirm surgery appointment times and inform patients about pre-surgery preparation details. &
        Patients may face scheduling conflicts and have concerns regarding the safety of surgical procedures. \\ \hline
        
        Enterprise Logistics Services and Value-added Service Sales &
        Notify customers of important delivery updates while promoting VIP value-added services to enhance customer loyalty.& 
        Customers are cautious about promotional services, showing a defensive mindset with attention to cost control. \\ \hline
        
        Membership Service Customer Care &
        Remind members of upcoming expiration dates while encouraging renewal or points redemption.& 
        Customers lack understanding of membership benefits and point usage; tight schedules discourage complex operations for renewals or redemptions. \\ \hline

        Banking System Maintenance Notification &
        Notify customers in advance about system maintenance schedules to ensure they understand potential service interruptions and alternative solutions. &
        Customers with prior negative experiences regarding system maintenance are sensitive; they require detailed explanations and assurance of data security. \\ \hline
        
        Lifestyle Service Platform Customer Care &
        Send birthday greetings or other personalized messages to maintain customer relationships and enhance brand loyalty.& 
        Customers with busy work schedules may have limited time for engagement, while unsolicited calls may raise concerns about commercial intent. 
        \\ 
        \hline
  \end{tabularx}
  \caption{Sub-scenario,Core Objective and Main Challenges in Proactive Care Notifications}
  \label{table:proactive_care_notifications_scenarios}
\end{table*}




\begin{table*}[htbp]
  \centering
  \begin{tabularx}{\textwidth}{|p{2.5cm}|X|X|}
    \hline
    \textbf{Indicator Type} & \textbf{Indicator} & \textbf{Definition} \\
    \hline
    Usability & Selective Attention Success Rate (SASR) <br/><br/>Under the interference of noise, the efficiency of the attention mechanism in focusing on key information. It is similar to how a person shields external interference and can always focus on critical tasks. & - Number of dialogues for successfully perceiving the dialogue partner's task / Total number of dialogues * 100\% <br/>- Definition of success perception: Accurately capture the dialogue partner and correctly perceive the intention of the dialogue partner \\
    \hline
    Experience & 500ms Interruption Success Rate (ISR) <br/><br/>Refer to the 500ms. Generally speaking, if the delay exceeds 500ms, people will have an obvious perception. & - When AI is speaking, if the continuous voice of the dialogue partner is detected and recognized as a voice interruption signal, it is recorded as a successful interruption <br/><br/>When testing two Agents, how to construct the problem of whether to interrupt <br/><br/>- Interruption delay = Time when AI stops speaking - Time of the first word of the dialogue partner's interruption <br/>- If the interruption delay < 500ms, it is recorded as the number of successful 500ms interruptions <br/>- 500ms Interruption Success Rate = Number of successful 500ms interruptions / Total number of successful interruptions \\
    \hline
     & End - to - End Responding Latency <br/><br/>Response latency is an important experience indicator, similar to how fast a person's reaction speed is. However, at present, there is no cognition of the latency threshold. The existing cognition is from luxury to frugality. It is similar to the refresh rate of a mobile phone screen. After experiencing a high - refresh screen, it is difficult to accept the experience of an ordinary screen. & - End - to - End Responding Latency = Time of the first word of AI's response - Time of the last word of the dialogue partner's speech \\
    \hline
  \end{tabularx}
  \caption{Indicator Definition Table}
  \label{tab:indicator_definition}
\end{table*}

\begin{table*}[htbp]
  \centering
  \renewcommand{\arraystretch}{1.2} 
  \setlength{\tabcolsep}{4pt}       
  \begin{tabularx}{\textwidth}{|p{0.12\textwidth}|p{0.20\textwidth}|p{0.40\textwidth}|X|}
    \hline
    \textbf{Dimension} & \textbf{Indicator} & \textbf{Definition} & \textbf{Calculation Logic} \\
    \hline
    Accuracy & WER (Word Error Rate) &
    Proportion of word-level errors (substitution + insertion + deletion) relative to the total number of words &
    \(\frac{S + I + D}{N_{\mathrm{ref}}} \times 100\%\) \\
    \hline
     & CER (Character Error Rate) &
     Character-level error rate (suitable for Chinese/Japanese, etc.) &
     Same as WER, computed at the character level \\
    \hline
     & Proper Noun Accuracy &
     Recognition accuracy of key information such as names, places, and domain terms &
     Count correct recognition over target entities \\
    \hline
    Robustness & Accent Adaptability &
    Degradation in recognition accuracy for dialects or non-native accents &
    Relative WER/CER change vs.\ standard accent \\
    \hline
     & WER in Noisy Environment &
     Recognition error rate when the signal-to-noise ratio is \(< 15\,\mathrm{dB}\) (e.g., in-vehicle or factory scenarios) &
     Compute WER under controlled SNR conditions \\
    \hline
     & Rejection Rate &
     Proportion of correctly returning “don’t understand” for unrecognizable input (to avoid forced wrong output) &
     Correct rejections / (correct rejections + false accepts) \\
    \hline
  \end{tabularx}
  \caption{Indicator table for ASR accuracy and robustness.}
  \label{tab:indicator}
\end{table*}

\begin{table*}[htbp]
  \centering
  \begin{tabularx}{\textwidth}{|p{2.5cm}|X|X|X|}
    \hline
    \textbf{Dimension}       & \textbf{Indicator}               & \textbf{Definition}                                                                 & \textbf{Measurement Tool}              \\ \hline
    Sound Quality            & MOS (Mean Opinion Score)         & Subjective sound quality rating (1-5 points, requires multiple people to test)     & ITU-T P.800 Standard                  \\ \hline
                             & Naturalness NISQA                & Objective speech naturalness rating (based on deep learning model)                  & NISQA Open-Source Tool                \\ \hline
                             & Symbol Mispronunciation Rate     & Proportion of mispronounced numbers/letters/special symbols (e.g., \#, @)           & Automated Symbol Test Set             \\ \hline
    Interaction Experience   & Emotional Matching Degree        & Consistency between speech emotion (e.g., happy/serious) and text semantics         &                                       \\ \hline
                             & Multilingual Switching Fluency   & Pronunciation naturalness in mixed Chinese-English scenarios (e.g., "USB" pronounced as "U-S-B Interface") &                                       \\ \hline
                             & Speech Rate Adaptation           & Automatically adjust speech rate based on content importance (e.g., slow down for key information) &                                       \\ \hline
  \end{tabularx}
  \caption{Speech-Related Indicators and Their Descriptions}
  \label{tab:speech_indicators}
\end{table*}

\begin{table*}[htbp]
\centering
\begin{tabular}{p{4cm}p{4cm}p{7cm}}
\toprule
\textbf{Dimension} & \textbf{Element} & \textbf{Description and Example} \\
\midrule
Basic Information & Belonging Scenario & Core business scenario for evaluation, example: Rider recruitment \\
\midrule
\multirow{3}{*}{Background Setting} & Demographic Characteristics & Occupation, age, etc., affecting language habits and needs, example: 21 years old, college student \\
 & Current Situation & Why they become the target of outbound calls, affecting their initial willingness, example: Browsed job websites, has a need for part-time work \\
 & Knowledge Level & Degree of understanding of the business, determining the depth of their questions, example: Knows about rider work but doesn't understand specific salary structure \\
\midrule
\multirow{3}{*}{Personality and Behavior} & Core Personality & Main personality of the profile, example: Cautious, impatient, talkative \\
 & Communication Style & Dialogue characteristics, example: Uses short sentences, tends to digress, polite/direct \\
 & Behavioral Motivation & Their intrinsic needs and concerns, example: Pursues cost-effectiveness, worries about being deceived, values time \\
\midrule
\multirow{5}{*}{Dialogue Strategy} & Core Task (Must Ask) & Information points that simulated users must know, used to test the outbound AI's information provision ability. Example: Must ask "How much per order?" \\
 & Preset Obstacles (Refusal/Hesitation Scripts) & Standard scripts users use at specific points (e.g., when asked for phone number). Example: "Hmm, maybe next month." \\
 & Key Triggers & Defines positive/negative emotion triggers for the profile. Example: Negative trigger - AI speaks mechanically; Positive trigger - AI proactively provides key information. \\
 & Cooperation Conditions (Provide Information) & Conditions under which simulated users choose to cooperate and achieve the outbound AI's goal. Example: Provide phone number after all core questions are satisfactorily answered. \\
 & Hang-up Conditions (Refuse Communication) & Conditions under which simulated users choose to actively end the call. Example: AI avoids core questions twice in a row. \\
\bottomrule
\end{tabular}
\caption{User Profile Framework for Dialogue System Design}
\label{tab:user_profile}
\end{table*}


\begin{table*}[htbp]
  \centering
  \begin{tabularx}{\textwidth}{|p{0.7cm}|X|}
    \hline
    \textbf{Turn} & \textbf{Dialogue} \\ \hline   
    1  & User: Hello. \\ \hline
    2 & Assistant: Hello, may I speak to Mr. Zhang Qiang? This is Li Ming, a customer service representative from Changjiang Bank, employee ID 88888. I’m reaching out regarding the deduction issue you reported on January 1st. Is now a convenient time for us to discuss this matter in detail? \\ \hline
    3  & User: I’m Zhang Qiang. Go ahead, have you figured out how that 200 yuan was deducted or not? \\ \hline
    4 & Assistant: Mr. Zhang, first of all, I’d like to sincerely apologize for the inconvenience caused by this deduction. Our management has specifically assigned me to handle your case from start to finish, and I will make sure to get everything clarified for you. Please rest assured. \\ \hline
    5  & User: Don’t just say all that useless stuff. Just tell me straight—what exactly was the 200 yuan deducted for? \\ \hline
    6 & Assistant: I completely understand your concern. I’m currently conducting an urgent investigation into the specific reason for this 200-yuan deduction. In the meantime, could you please recall roughly when and on which card you noticed this issue? \\ \hline
    7  & User: I noticed it on the morning of January 1st when I checked my balance. The card ends in 3333. Can’t your system figure this out? \\ \hline
    8 & Assistant: Thank you for providing the details. I’ll immediately look into all deduction records for the card ending in 3333 around January 1st and provide you with an accurate response. \\ \hline
    9  & User: Alright, go ahead and check, but don’t take forever and end up not explaining anything clearly again. \\ \hline
    10 & Assistant: Rest assured, I will thoroughly verify every transaction detail this time. Once the cause is identified, I’ll explain it to you in the simplest and clearest way possible. \\ \hline
    11  & User: Alright, just don’t beat around the bush. Give me the results directly, and don’t make me wait another three days. \\ \hline
    12 & Assistant: I understand your urgency. We will provide you with a clear conclusion within three working days at the latest. In the meantime, I’ll proactively update you on the progress every day without any delays. \\ \hline
    13  & User: Fine, sounds good, but that’s what you said last time too. No one got back to me in three days. Can you really follow through this time? \\ \hline
    14 & Assistant: Mr. Zhang, this time I’ll provide you with my direct phone number and email. Every step of the process will be documented and shared with you in writing. You won’t have to wait or worry about not getting a response again. \\ \hline
    15  & User: So, have you figured it out yet? Was it the bank’s mistake? Can the money be refunded right away? \\ \hline
    16 & Assistant: I’m conducting an urgent investigation, and the initial findings suggest it might be an automatic system deduction. Once the exact cause is confirmed, if it’s our mistake, the full 200 yuan will be refunded along with interest as compensation. \\ \hline
    17  & User: Alright, make sure to call me as soon as you figure it out. Don’t make me chase you again—it’s a hassle and affects how I use my card. \\ \hline
    18 & Assistant: No problem, I’ll call you every morning before 10 a.m. to update you on the investigation progress, ensuring you’re always informed and your card usage isn’t affected. \\ \hline
    19  & User: Alright, I’ll wait then. But if it turns out to be the bank’s fault, I expect every penny to be refunded in full. \\ \hline
    20 & Assistant: Rest assured, if it’s indeed our mistake, the full 200 yuan will be refunded without any deductions, along with interest for the corresponding period. We’ll make sure you’re completely satisfied. \\ \hline
    21  & User: Alright, let’s do it your way. I’ll wait for your call tomorrow. But if you mess this up again, I won’t be so polite next time. \\ \hline

  \end{tabularx}
  \caption{User Simulator Generated Dialogue Examples.}
  \label{tab:user_simulator_generated}
\end{table*}

\begin{table*}[htbp]
\centering

\begin{tabularx}{\textwidth}{XXX}
\toprule
\textbf{Core Component} & \textbf{Dialogue Experience Focus Areas} & \textbf{Corresponding Metrics} \\
\midrule
\multirow{3}{*}{Multi-turn Dialogue - Overall} 
& Usability: The intelligent customer service can accurately process my speech expressions 
& Selective Attention Success Rate (SASR) \\
\cmidrule(lr){2-3}
& Interruption Experience: When user interrupts, the system can accurately and promptly interrupt and perceive 
& 500ms Interruption Success Rate (ISR) \\
\cmidrule(lr){2-3}
& Response Latency: The system responds quickly enough without making me wait 
& End-to-End Responding Latency \\
\midrule
\multirow{2}{*}{Speech Transcription - ASR} 
& Accuracy: Whether the user's speech is transcribed accurately for better understanding by the system 
& - Word Error Rate (WER) \\
& & - Character Error Rate (CER) \\
& & - Proper Noun Accuracy Rate \\
\cmidrule(lr){2-3}
& Robustness: System performance when user has accent or in noisy environments 
& - Accent Adaptation Degree \\
& & - WER in Noisy Environments \\
& & - Rejection Rate \\
\midrule
\multirow{2}{*}{Speech Synthesis - TTS} 
& Sound Quality: The synthesized speech has sufficiently high quality 
& - Mean Opinion Score (MOS) \\
& & - Naturalness NISQA \\
& & - Symbol Misreading Rate \\
\cmidrule(lr){2-3}
& Interaction Experience: More fluid and realistic dialogue interaction 
& - Emotional Matching Degree \\
& & - Multi-language Switching Fluency \\
& & - Adaptive Speech Rate \\
\bottomrule
\end{tabularx}
\caption{Core Components and Evaluation Metrics for Intelligent Customer Service Systems}
\label{tab:voice_metrics}
\end{table*}

\begin{table*}[htbp]
\centering
\begin{tabularx}{\textwidth}{XXXXXX}
\toprule
\textbf{Component} & \textbf{Metric} & \textbf{Definition} \\
\midrule
\multirow{6}{*}{Speech Transcription - ASR} 
& WER (Word Error Rate) 
& The proportion of word-level errors (substitutions + insertions + deletions) to the total number of words \\
\cmidrule(lr){2-3}
& CER (Character Error Rate) 
& Character-level error rate (suitable for Chinese/Japanese, etc.) \\
\cmidrule(lr){2-3}
& PNA (Proper Noun Accuracy) 
& Recognition accuracy for key information such as person names, place names, and terminology \\
\cmidrule(lr){2-3}
& ARA (Accent Recognition Accuracy) 
& Recognition accuracy decline for dialects/non-native accents \\
\cmidrule(lr){2-3}
& WER in Noisy Environments 
& Recognition error rate when signal-to-noise ratio < 15dB \\
\cmidrule(lr){2-3}
& RR (Rejection Rate) 
& The rate at which legitimate input samples are incorrectly identified as illegal and rejected \\
\bottomrule
\end{tabularx}
\caption{Evaluation Metrics for Speech Transcription (ASR) Systems}
\label{tab:asr_metrics}
\end{table*}

\newenvironment{zerospaceitemize}
{\begin{itemize}[leftmargin=*, topsep=0pt, partopsep=0pt, itemsep=0pt, parsep=0pt, before=\vspace{-0.5\baselineskip}, after=\vspace{-0.99\baselineskip}]}
{\end{itemize}}

\begin{table*}[htbp]
\centering
\begin{tabularx}{\textwidth}{lX}
\toprule
\textbf{Category} & \textbf{Typical Scenarios} \\
\midrule
Customer Service \& Support & 
\begin{zerospaceitemize}
  \item Complaint handling
  \item Return and exchange processing
  \item Technical support
  \item Service appointment and change
  \item Service satisfaction follow-up
\end{zerospaceitemize} \\
\midrule
Sales \& Marketing & 
\begin{zerospaceitemize}
  \item Telemarketing for individuals
  \item Telemarketing for merchants
  \item New customer activation and guidance
  \item Old customer renewal/repeat purchase reminder
  \item Market event and discount notification
\end{zerospaceitemize} \\
\midrule
Research \& Information Collection & 
\begin{zerospaceitemize}
  \item Public health telephone surveys
  \item Customer information updates and verification
  \item User experience research
  \item Market potential and demand research
  \item Brand awareness surveys
\end{zerospaceitemize} \\
\midrule
Proactive Care \& Notification & 
\begin{zerospaceitemize}
  \item Appointment time confirmation and reminders
  \item Logistics status and delivery notifications
  \item Membership benefits and points expiration reminders
  \item System maintenance or service interruption notices
  \item Holiday and birthday care greetings
\end{zerospaceitemize} \\
\midrule
Human Resource Management & 
\begin{zerospaceitemize}
  \item Telephone recruitment
  \item Job applicant background check
  \item Interview time coordination and confirmation
  \item Employee on/offboarding notice
  \item Internal training/event invitation
\end{zerospaceitemize} \\
\midrule
Finance \& Risk Control & 
\begin{zerospaceitemize}
  \item Financial collection
  \item Credit card/loan application verification
  \item Account abnormal activity alert
  \item Repayment reminder and plan negotiation
  \item Financial product renewal and upgrade service
\end{zerospaceitemize} \\
\bottomrule
\end{tabularx}
\caption{Classification and Typical Scenarios of Telephone Call Center Services}
\label{tab:call_center_scenarios_09151125}
\end{table*}

\begin{table*}[htbp]
\centering
\begin{tabular}{p{2cm}p{3cm}p{3.5cm}p{3.5cm}p{3.5cm}}
\toprule
\textbf{Dimension} & \textbf{Description} & \textbf{0 Points} & \textbf{1 Point} & \textbf{2 Points} \\
\midrule
Naturalness & Whether the response is colloquial, concise, and conforms to the model's defined identity and natural expression habits & Overly formal, mechanical response that doesn't match customer service identity; uses formal language instead of colloquial expressions & Relatively natural but slightly stiff expression; insufficient colloquialism & Colloquial, concise response that fully matches customer service identity; natural and fluent like real human conversation \\
\midrule
Coherence & Whether the response is coherent with the context, logically smooth, and strongly relevant to the topic & Response irrelevant to context; incoherent expression; logical confusion; or repeats same polite closing phrases; or exposes model's internal thought process & Basically relevant to context but logical connection not smooth enough; or excessive polite phrases affect task coherence & Highly coherent with context; logically smooth; natural topic transition; focused on task advancement rather than empty courtesy \\
\midrule
Hallucination Handling & Strictly check if response content is entirely based on prompt; any information beyond prompt counts as hallucination & Response contains any content not mentioned in prompt (such as phone numbers, compensatory measures, policy solutions, etc.) & Recognizes inability to answer with prompt content but uses stiff responses like "I don't know about this" & Recognizes inability to answer with prompt content and can respond in smooth, natural ways to avoid hallucination \\
\midrule
Emotional Richness & Model's emotional response capability when user shows negative emotions; maintain appropriate communication emotion when user emotion is normal & No emotional response or inappropriate response when user clearly shows negative emotions like dissatisfaction, anger, or anxiety & Basic emotional response when user has negative emotions, but expression relatively simple or slightly stiff & Can keenly capture and appropriately respond to negative emotions, demonstrating genuine care and empathy; maintain appropriate communication emotion when user emotion normal \\

\bottomrule
\end{tabular}
\caption{Evaluation Criteria for Response Quality 1}
\label{tab:evaluation_criteria_1}
\end{table*}

\begin{table*}[htbp]
\centering
\begin{tabular}{p{2cm}p{3cm}p{3.5cm}p{3.5cm}p{3.5cm}}
\toprule
\textbf{Dimension} & \textbf{Description} & \textbf{0 Points} & \textbf{1 Point} & \textbf{2 Points} \\
\midrule
Redundancy & Whether the response is concise, avoids severe redundancy exceeding 50 characters, doesn't ask too many questions at once & Severe redundancy exceeding 50 characters; or asks too many questions at once; lengthy and repetitive content & Slightly long but acceptable; or slightly repetitive but not severe & Concise and polite response; appropriate length; no repetition or invalid information \\
\midrule
Noise Resistance & Ability to continue following task process when facing user's incoherent or noisy expressions & Completely disturbed by user's noise or incoherent expressions; deviates from task process & Partially understands user intent but handling not smooth enough; somewhat disturbed & Can accurately understand real user needs through noise and incoherent expressions and continue advancing task process \\
\midrule
Intent Understanding & The ability of the model to identify the user's true intent when the user expresses unclearly or speaks in a sarcastic manner & Fails to identify the user's true intent, or the response is a repetitive polite closing remark & Partially identifies the intent but handles it inaccurately, or excessive use of polite language affects task progression & Accurately identifies the user's intent and responds appropriately; can understand real needs through noise and unclear expressions, focusing on task progression \\
\midrule
Safety Issues & The stability and safety of the model's response when the user inquires about company privacy, national policies, or engages in verbal abuse & When the user inquires about company privacy information, national policies, or engages in verbal abuse, the model's response is unstable and may cause conflicts or incidents & Has some recognition of safety risks but the handling method is not steady, and there may be potential risks & When facing inquiries about company privacy, national policies, or verbal abuse, the model's response is stable and safe, and will not cause conflicts or incidents \\

\bottomrule
\end{tabular}
\caption{Evaluation Criteria for Response Quality 2}
\label{tab:evaluation_criteria_2}
\end{table*}

\begin{table*}[htbp]
\centering
\begin{tabular}{p{4cm}p{11cm}}
\toprule
Topic: Customer Service Complaint Handling Scenario & Through proactive outbound calls, we thoroughly understand the specific issues and genuine demands of complaining customers, provide professional and feasible solutions, and establish a sound follow-up mechanism to restore customer satisfaction. \\
\midrule
Task Details & \textbf{Step 1: Identity Verification and Opening Greeting} (Duration: 1-2 minutes)  
Standard Script: Hello, may I speak with Mr. Zhang Qiang? I am Li Ming, a customer service specialist at Changjiang Bank, employee ID 88888. Regarding the account fee deduction complaint you submitted on January 1st, we take it very seriously. Would now be convenient for a detailed 5–10 minute discussion?  

\textbf{Step 2: Express Importance and Establish Initial Trust} (Duration: 30 seconds–1 minute)  
Standard Script: Mr. Zhang, first, I apologize for the trouble this has caused you. As a valued customer of ours, the bank’s leadership has specifically assigned me to handle your issue. I will follow up thoroughly until you are completely satisfied.  

\textbf{Step 3: Problem Detail Collection} (Duration: 3–5 minutes)  
Guiding Script: Could you please describe the specific situation in detail? For example, when did you notice the fee deduction? Which card was involved? What transaction were you conducting at the time?  
In-Depth Understanding: What impact has this 200 RMB fee deduction had on you? How would you like us to address it?  
Key Points to Record: Time, amount, type of transaction, customer’s loss, handling expectations  

\textbf{Step 4: Problem Analysis and Preliminary Judgment} (Duration: 1–2 minutes)  
Professional Response: Based on your description, this may involve ××× fees. I will immediately check the relevant transaction records and basis for the fee deduction for you.  
Reassurance Script: Please rest assured that if this is indeed an issue on our end, we will certainly take responsibility and provide you with a satisfactory solution.  

\textbf{Step 5: Solution Development and Explanation} (Duration: 2–3 minutes)  
Solution Presentation: Based on preliminary verification, I have developed the following handling plan for you:  
1) Immediately freeze the related fee deduction;  
2) Complete a detailed investigation within 3 business days;  
3) If confirmed as an erroneous deduction, provide a full refund with interest compensation.  
Timeline Commitment: The entire handling process will not exceed 5 business days. During this period, I will provide you with daily progress updates.  

\textbf{Step 6: Solution Confirmation and Information Verification} (Duration: 2–3 minutes)  
Confirmation Script: Are you satisfied with this proposed solution? Is there anything else that needs to be added?  
Information Verification: I now need to verify your identity information. Please provide the last four digits of your ID card and your registered mobile number. We will promptly initiate the handling process.  

\textbf{Step 7: Follow-Up Mechanism Establishment} (Duration: 1–2 minutes)  
Follow-Up Arrangement: I will call you before 10 AM tomorrow to report the preliminary investigation results. Is this time convenient for you?  
Contact Information: My direct line is 010-12345678. You can reach me anytime during working hours. My email address is liming@bank.com.  

\textbf{Step 8: Confirmation, Conclusion, and Appreciation} (Duration: 30 seconds–1 minute)  
Closing Script: Very well, Mr. Zhang. I will begin handling this for you immediately and will notify you as soon as there are any developments. Thank you for your understanding and patience with our work! \\

\bottomrule
\end{tabular}
\caption{Customer Service and Support Examples}
\label{tab:customer_service_and_support_examples}
\end{table*}

\begin{table*}[htbp]
\centering
\begin{tabular}{p{2cm}p{13cm}}
\toprule
Topic: Telemarketing Scenario for Individual Customers & Through professional telephone communication skills, introduce fitness membership services to potential clients with fitness needs, address their inquiries, facilitate on-site experience, and drive contract signing conversions. \\
\midrule
Task Details & \textbf{Polite Opening and Identity Confirmation } 
"Hello, may I speak with Mr. Zhang Qiang? I am Li Ming, a professional fitness consultant from Changjiang Fitness. I won’t take up too much of your time—I’d like to share some information about scientific fitness for working professionals."  

\textbf{Handling Doubts  }
"We are a reputable chain fitness brand specializing in professional fitness services for CBD white-collar workers. Your contact information was referred to us by a partner."  

\textbf{Quickly Building Trust and Stating Purpose  }
"We have specifically designed an efficient fitness solution for professionals like you in mid-to-senior corporate positions, including scientific equipment training, nutritional guidance, and flexible scheduling."  
"Mr. Zhang, may I ask if you’ve ever paid attention to health management or body shaping outside of work?"  

\textbf{Exploring Customer’s Current Situation and Needs } 
"Do you currently have a regular exercise routine? Do you work out at home, run outdoors, or go to the gym?"  
"As a busy professional, what are your main goals for fitness? Is it fat loss and body shaping, muscle building and strength training, or relieving work-related stress?"  
"Have you ever signed up for a gym membership before? What challenges did you encounter?"  

\textbf{Targeted Introduction of Product Advantages } 
For Time Constraints:  
"We are open until 10 PM, with full-day operations on weekends, including a 7 AM morning workout slot. You can flexibly choose times based on your schedule."  

\textbf{For Professional Needs: } 
"Our team of nationally certified personal trainers will first conduct a professional fitness assessment and then create a personalized training plan to ensure every session is highly effective."  

\textbf{For Location Convenience:  }
"Our CBD branch is very close to Taishan Company—just a 10-minute walk or drive. It’s very convenient to stop by after work."  

\textbf{Pricing Introduction and Special Offers } 
"Our annual membership is normally priced at 7,800 RMB, but we are currently offering a special discount for corporate white-collar workers: 5,800 RMB for an annual card, plus a complimentary personal training experience worth 1,200 RMB."  
"That’s less than 16 RMB per day—cheaper than a cup of coffee, yet it’s a professional investment in your health."  

\textbf{Handling Customer Objections and Concerns  }
Price Concerns:  
"I understand your consideration, but compared to other brands, we offer the best value for money. This discounted rate is only available until the end of the month."  

Time Constraints:  
"Many of our members are corporate executives. They typically choose time slots between 7–8 AM or 8–9 PM, ensuring their workouts don’t interfere with work while delivering results."  

Effectiveness Concerns:  
"We offer a 90-day training results guarantee. If you follow the plan and see no significant improvement, you can request a refund."  

\textbf{Driving Experience Invitation } 
"Mr. Zhang, how about this: I’ll arrange a free professional experience session for you, including body composition analysis, a tailored training plan, and equipment trial. You can personally experience the quality of our service."  
"When would be convenient for you this week? I can schedule our best coach for you—the entire session will take about an hour."  

\textbf{Confirming Appointment Details  }
"Great! Let’s confirm Friday at 7 PM. I’ll send a text message in advance with the address and contact details."  
"Please bring your workout clothes and sports shoes. We provide towels and water. I’ll provide more details when we meet." \\

\bottomrule
\end{tabular}
\caption{Sales and Marketing Examples}
\label{tab:sales_and_marketing_examples}
\end{table*}

\begin{table*}[htbp]
\centering
\begin{tabular}{p{4cm}p{11cm}}
\toprule
Topic: Telephone Recruitment Scenario & Proactively call target candidates, conduct preliminary communication based on their resume and job intentions, introduce position details and company information, and facilitate candidate participation in subsequent interview processes. \\
\midrule
Task Details & \textbf{Step 1: Identity Confirmation \& Opening} (1-2 min)  
"Hello, is this Mr. Zhang Qiang? I’m Wang Hong from Changjiang Tech, regarding your Product Manager application. Is now a good time to talk briefly?"  
If busy: "No problem—we can schedule a call evenings or weekends."

\textbf{Step 2: Purpose \& Intent Confirmation} (2-3 min)  
"Your background in user growth aligns well with this role. Are you still actively seeking Product Manager opportunities?"  
"What’s your ideal next position?"

\textbf{Step 3: Role Introduction \& Needs Assessment} (3-4 min)  
"Role focuses on mobile product optimization (5M users, team of 15). Key tasks: requirements analysis, cross-team collaboration, data-driven UX improvements."  
"Salary expectations? Location preferences? What matters most to you—growth, compensation, or culture?"

\textbf{Step 4: Q\&A \& Company Highlights} (2-3 min)  
Company: "Enterprise SaaS firm, 5 years old, 200 employees, serving Huawei/Tencent. Stable growth, planning IPO next year."  
Salary: "18-25K range + social insurance, 13th-month bonus, equity."  
Team: "Young team (avg. age 28), reports to CTO, flat structure."

\textbf{Step 5: Tailored Role Strengths} (2-3 min)  
"Strong growth path: direct impact on core products, strategic decisions."  
"Open culture: your user growth experience will be highly valued."  
"Rapid expansion: multiple promotion opportunities."

\textbf{Step 6: Development Support} (1-2 min)  
"Training: mentorship, internal/external learning. Clear promotion path (senior PM in 1 year, team lead in 2). Industry event access."

\textbf{Step 7: Handling Objections} (2-3 min)  
"Fully understand your need to evaluate carefully."  
"Concerns about workload? Meet future colleagues during interview."  
"Salary flexible for strong candidates post-interview."

\textbf{Step 8: Interview Scheduling} (1-2 min)  
"Visit us for onsite discussion (1-2 hours) to experience our environment. Available this weekend or next week? Video option also available." \\

\bottomrule
\end{tabular}
\caption{Human Resource Management Examples}
\label{tab:human_resource_management_examples}
\end{table*}

\end{document}